\documentclass[10pt,twocolumn,letterpaper]{article}

\usepackage{cvpr}
\usepackage{times}
\usepackage{epsfig}
\usepackage{graphicx}
\usepackage{amsmath}
\usepackage{amssymb}

\usepackage{bbm}
\usepackage{breqn}
\usepackage[noend, boxruled, linesnumbered] {algorithm2e}
\usepackage[title]{appendix}

\DeclareMathOperator*{\argmax}{arg\,max}
\DeclareMathOperator*{\argmin}{arg\,min}

\usepackage[pagebackref=true,breaklinks=true,letterpaper=true,colorlinks,bookmarks=false]{hyperref}

\cvprfinalcopy 

\def\cvprPaperID{4837} 

\ifcvprfinal\fi
\begin{document}

\title{Dissimilarity Coefficient based Weakly Supervised Object Detection}

\author{Aditya Arun\\
IIIT Hyderabad\\
\and
C.V. Jawahar\\
IIIT Hyderabad
\and
M. Pawan Kumar\\
University of Oxford,\\ 
The Alan Turing Institute
}

\maketitle

\begin{abstract}
We consider the problem of weakly supervised object detection, where the training samples are annotated using only image-level labels that indicate the presence or absence of an object category. In order to model the uncertainty in the location of the objects, we employ a dissimilarity coefficient based probabilistic learning objective. The learning objective minimizes the difference between an annotation agnostic prediction distribution and an annotation aware conditional distribution. The main computational challenge is the complex nature of the conditional distribution, which consists of terms over hundreds or thousands of variables. The complexity of the conditional distribution rules out the possibility of explicitly modeling it. Instead, we exploit the fact that deep learning frameworks rely on stochastic optimization. This allows us to use a state of the art discrete generative model that can provide annotation consistent samples from the conditional distribution. Extensive experiments on PASCAL VOC 2007 and 2012 data sets demonstrate the efficacy of our proposed approach.
\end{abstract}

\vspace{-3mm} 
\section{Introduction} \label{sec1:intro}
Object detection requires us to localize all the instances of an object category of interest in a given image. In recent years, significant advances in speed and accuracy have been achieved by detection frameworks based on Convolutional Neural Networks (CNNs) \cite{dai2016r,girshick15fastrcnn,girshick2014rcnn,he2017mask,liu2016ssd,Redmon_2016_CVPR,ren2015faster}. Most of the existing methods require a strongly supervised data set, where each image is labeled with the ground-truth bounding boxes of all the object instances. Given the high cost of obtaining such detailed annotations, researchers have recently started exploring the weakly supervised object detection (WSOD) problem~\cite{Bilen_2016_CVPR,Diba_2017_CVPR,Jie_2017_CVPR,lai2017saliency,Li_2016_CVPR,li2017multiple,Tang_2017_CVPR,Tang_2018_ECCV,yan2017weakly,Zhang_2018_CVPR,Zhang_2018_ECCV,Zhang_2018_CVPR_W2F}. The goal of WSOD is to learn an accurate detector using training samples that are annotated with image-level labels (which indicate the presence of an object category).

Given the wide availability of image-level labels, WSOD offers a cost-effective and highly scalable learning paradigm. However, this comes at the cost of introducing uncertainty in the location of the object instances during training. For example, consider the task of detecting a car. Given a training image annotated to indicate the presence of a car, we are still faced with the challenge of identifying the bounding box for the car. 

In order to effectively model uncertainty in weakly supervised learning, Kumar \emph{et al.}~\cite{kumar2012modeling} proposed a probabilistic framework that models two distributions: (i) a conditional distribution, which represents the probability of an output conditioned on the given annotation during training; and (ii) a prediction distribution which represents the probability of an output at test time. The parameters of the two distributions are estimated jointly by minimizing the dissimilarity coefficient~\cite{rao1982diversity}, which measures the distance between any two distributions using a task specific loss function.

The aforementioned dissimilarity coefficient based framework has provided promising results in domains where the conditional distribution is simple to model (that is, consists of terms that depend on a few variables at a time)~\cite{arun2018learning, kumar2012modeling}. However, WSOD presents a more challenging scenario due to the complexity of the underlying conditional distribution. Specifically, given the hundreds or even thousands of bounding box proposals for an image, the annotation constraint imposes a term over all of these bounding box proposals such that at least one of them corresponds to the given image-level label. This leads to a challenging scenario where the distribution is not factorizable over the bounding box proposals. While previous works have approximated this uncertainty as a fully factorized distribution for computational efficiency, we argue that such a choice leads to poor accuracy. 

To overcome the difficulty of a complex conditional distribution, we make the key observation that 
deep learning relies on stochastic optimization. Therefore, we do not need to explicitly model this complex distribution but simply estimate the distribution using samples. This observation opens the door to the use of state-of-the-art deep generative models such as the Discrete \textsc{Disco} Net~\cite{bouchacourt2017thesis}. 

We test the efficacy of our approach on the challenging PASCAL VOC 2007 and 2012 data sets. To generate the weakly supervised data sets, we use the image-level labels, discarding the bounding box annotations. We achieve $53.6\%$ detection AP on PASCAL VOC 2007 and $49.5\%$ detection AP on PASCAL VOC 2012 data set, significantly improving the state-of-the-art by $1.5\%$ on both data sets.

To summarize, we make the following contributions.
\begin{itemize}
\itemsep=-0.25em
    \item Efficiently model the complex non-factorizable, annotation aware conditional distribution using the deep generative model, the Discrete \textsc{Disco} Net.
    \item Empirically show the importance of modeling the uncertainty in the annotations in a single unified probabilistic learning objective, the dissimilarity coefficient.
    \item State-of-the art performance for the task of WSOD on challenging PASCAL VOC 2007 and 2012 data sets.
\end{itemize}

\section{Related Work} \label{sec2:relWork}
Conventional methods often treat WSOD as a Multiple Instance Learning (MIL) problem~\cite{dietterich1997solving} by representing each image as a bag of instances (that is, putative bounding boxes)~\cite{bilen2015weakly,cinbis2017weakly,song2014weakly,wang2014weakly,wang2015relaxed}. The learning procedure alternates between training an object classifier and selecting the most confident positive instances. However, these methods are susceptible to poor initialization. To address this, different strategies have been developed, which aim to improve the initialization~\cite{kumar2010self,siva2012defence,siva2011weakly,song2014weakly}, regularize the model with extra cues~\cite{bilen2015weakly,cinbis2017weakly}, or relax the MIL constraint~\cite{wang2015relaxed} to make the objective differentiable. These hard-MIL based methods have demonstrated their effectiveness, specially when CNN features are used to represent object proposals~\cite{cinbis2017weakly}. However, these models are not end to end trainable and also do not explicitly model the uncertainty. 

A more interesting line of work is to integrate MIL strategy as deep networks such that they are end to end trainable~\cite{Bilen_2016_CVPR,Diba_2017_CVPR,Ge_2018_CVPR,Tang_2017_CVPR,Tang_2018_ECCV,wang2018collaborative,Zhang_2018_CVPR,Zhang_2018_ECCV,Zhang_2018_CVPR_W2F}. In their work, Bilen \emph{et al.}~\cite{Bilen_2016_CVPR} proposed a smoothed version of MIL that softly labels object proposals instead of choosing the highest scoring ones. Building on this soft-MIL based approach, Diba \emph{et al.}~\cite{Diba_2017_CVPR} integrate the MIL strategy with better bounding box proposals into an end-to-end cascaded deep network. Tang \emph{et al.}~\cite{Tang_2017_CVPR} refine the prediction iteratively through multi-stage instance classifier. Zhang \emph{et al.}~\cite{Zhang_2018_CVPR} add curriculum learning using the MIL framework. As we shall see, our formulation brings out the curriculum learning naturally during training. Other end-to-end trainable frameworks for WSOD employ domain adaptation~\cite{Li_2016_CVPR,song2014weakly}, expectation-maximization algorithm~\cite{Jie_2017_CVPR,yan2017weakly} and saliency based methods~\cite{lai2017saliency}. Although these methods are end to end trainable, they not only model a single distribution for two related tasks, but also model the complex distribution with a fully factorized one. This makes these approach sub-optimal as what we truly want is to model a distribution which enforces at least one bounding box proposals corresponding to the image-level label.

There have been attempts to further improve the performance of the weakly supervised detectors by combining them with the strongly supervised detectors. Typically, the predicted instances from a trained weakly supervised detector are treated as a pseudo-strong label to train a strongly supervised network~\cite{Ge_2018_CVPR,Li_2016_CVPR,Tang_2017_CVPR,Tang_2018_ECCV,Zhang_2018_CVPR,Zhang_2018_ECCV,Zhang_2018_CVPR_W2F}. However, there is only a unidirectional connection between the two detectors. In their work, Wang \emph{ et al.} \cite{wang2018collaborative}  train a weakly and strongly supervised model jointly, in a collaborative manner. This is similar in spirit to ours in using two distributions. However, they model their weakly supervised detector with a fully factorized distribution. The improvement in results reported by these papers advocates the importance of modeling two separate distributions. In this work, we explicitly define the two distributions employed during training and test time and jointly train them by minimizing the dissimilarity coefficient~\cite{rao1982diversity} based objective function.

\section{Model} \label{sec3:model}
\subsection{Notation} \label{ssec3.1:notation}
We denote an input image as ${\bf x} \in \mathbb{R}^{(H \times W \times 3)}$, where $H$ and $W$ are the height and the width of the image respectively. For the sake of simplifying the subsequent description of our approach, we assume that we have extracted $B$ bounding box proposals from each image. In our experiments, we use Selective Search~\cite{uijlings2013selective}. Each bounding box proposal can belong to one of $C+1$ categories from the set $\{0,1,\dots,C\}$, where category $0$ is background, and categories $\{1,\dots,C\}$ are object classes. 

We denote an image-level label by ${\bf a} \in \{0, 1\}^C$, where ${\bf a}^{(j)} = 1$ if image ${\bf x}$ contains the $j$-th object.
Furthermore, we denote the unknown bounding box labels by ${\bf y} \in \{0, \dots, C\}^{B}$, where ${\bf y}^{(i)} = j$ if the $i$-th bounding box is of the $j$-th category. A weakly supervised data set $\mathcal{W} = \{({\bf x}_i, {\bf a}_i) | i = 1, \dots, N\}$ contains $N$ pairs of images ${\bf x}_i$ and their corresponding image-level labels ${\bf a}_i$. 

\subsection{Probabilistic Modeling} \label{ssec3.2:probModel}
Given a weakly supervised data set $\mathcal{W}$, we wish to learn an object detector that can predict the bounding box labels ${\bf y}$ of a previously unseen image. Due to the uncertainty inherent in this task, we advocate the use of a probabilistic formulation. Following \cite{arun2018learning,kumar2012modeling}, we define two distributions. The first one is the prediction distribution $\Pr_{p}({\bf y} | {\bf x};\boldsymbol{\theta}_p)$, which models the probability of the bounding box labels ${\bf y}$ given an input image ${\bf x}$. Here $\boldsymbol{\theta}_p$ are the parameters of the distribution. As the name suggest, this distribution is used to make the prediction at test time. 

In addition to the prediction distribution, we also construct a conditional distribution $\Pr_{c}({\bf y}| {\bf x}, {\bf a}; \boldsymbol{\theta}_{c})$, which models the probability of the bounding box labels ${\bf y}$ given the input image ${\bf x}$ and its image-level annotations ${\bf a}$. Here $\boldsymbol{\theta}_c$ are the parameters of the distribution. The conditional distribution contains additional information, namely the presence of foreground objects in each image. Thus, we can expect it to provide better predictions for the bounding box labels ${\bf y}$. We will use this property during training in order to learn an accurate prediction distribution using the conditional distribution. The details on the modeling of the two distributions are discussed below.

\begin{figure*}[h!]
\begin{center}
   \includegraphics[width=0.99\textwidth]{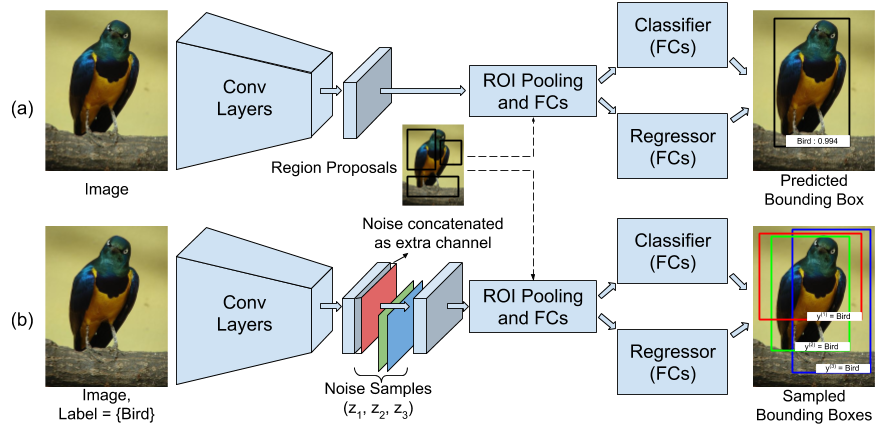} 	
\end{center}
   \caption{\emph{The overall architecture. (a) Prediction Network: a standard Fast-RCNN architecture is used to model the prediction net. For an input image, bounding box proposals are generated from selective search~\cite{uijlings2013selective}. Features from each of these proposals are computed by the region of interest (ROI) pooling layers, which are then passed through the classifier and regressor to predict the final bounding box. (b) Conditional Network: a modified Fast-RCNN architecture is used to model the conditional net. For a single input image ${\bf x}$ and three different noise samples $\{{\bf z}_1, {\bf z}_2, {\bf z}_3\}$ (represented as red, green and blue matrix), three different bounding boxes $\{{\bf y}^{(1)}, {\bf y}^{(2)}, {\bf y}^{(3)}\}$ are sampled for the given image-level label (bird in this example). Here the noise filter is concatenated as an extra channel to the final convolutional layer. For both the networks, the initial conv-layers are fixed during training. Best viewed in color.}}
\label{fig1:Arch}
\end{figure*}

\vspace{-2.5mm} 
\subsubsection{Prediction Distribution} \label{sssec3.2.1:predDist}
The task of the prediction distribution is to accurately model the probability of the bounding box labels given the input image. Taking inspiration from the supervised models \cite{girshick15fastrcnn,girshick2014rcnn,ren2015faster}, we assume independence between the probability of the output for each bounding box proposal. Therefore, the overall distribution for an image equals the product of the probabilities of each proposal,
\begin{equation} 
\label{eq1:pred_dist}
	\Pr\nolimits_{p}({\bf y}|{\bf x}; \boldsymbol{\theta}_{p}) = \prod_{i = 1}^{B} \Pr\nolimits_{p}({\bf y}^{(i)}|{\bf x}; \boldsymbol{\theta}_{p}).
\end{equation}
We model this distribution using the Fast-RCNN architecture~\cite{girshick15fastrcnn} (see Figure \ref{fig1:Arch}(a)). As the prediction distribution is specified by a neural network, we henceforth refer to it as the {\em prediction net}. In this setting, the parameters of the distribution $\boldsymbol{\theta}_p$ are the weights of the prediction net.

\vspace{-2.5mm} 
\subsubsection{Conditional Distribution} \label{sssec3.2.2:condDist}
Given $B$ bounding box proposals for an image ${\bf x}$ and the image-level label ${\bf a}$, the conditional distribution $\Pr_{c}({\bf y}| {\bf x, a}; \boldsymbol{\theta}_{c})$ models the probability of bounding box labels ${\bf y}$ under the constraint that they are compatible with the annotation ${\bf a}$. Specifically, there exists at least one bounding box $i$ such that ${\bf y}^{(i)} = j$, for every positive image-level label ${\bf a}^{(j)} = 1$.

Note that due to the requirement that the bounding box labels ${\bf y}$ are compatible with the annotation ${\bf a}$, the conditional distribution cannot be trivially decomposed over bounding box proposals. This is in stark contrast to the simple prediction net, which uses a fully factorized distribution. If one were to explicitly model the conditional distribution, then one would be required to compute its partition function during training, which would be prohibitively expensive. To alleviate this computational challenge, we make a key observation that in practice we only need access to a representative set of samples from the conditional distribution. This opens the door to the use of the recently proposed Discrete \textsc{Disco} Net~\cite{bouchacourt2017thesis}. In what follows, we briefly describe Discrete {\sc Disco} Nets while highlighting their applicability to our framework.

\vspace{-2.5mm} 
\paragraph{Discrete \textsc{Disco} Net:} Discrete \textsc{Disco} Net~\cite{bouchacourt2017thesis} is a deep probabilistic framework that implicitly represents a  probability distribution over a discrete structured output space. The strength of the framework lies in the fact that it allows us to adapt a pointwise deep network (a network that provides a single pointwise prediction) to a probabilistic one by the introduction of noise.

In the context of our setting, consider the modified Fast-RCNN network in Figure \ref{fig1:Arch}(b) for the conditional distribution. Once again, as we are using a neural network, we will henceforth refer to it as the \emph{conditional net}. The parameters of the conditional distribution $\boldsymbol{\theta}_c$ are the weights of the conditional net. The colored filters in the middle of the network represent the noise that is sampled from a uniform distribution. Each value of the noise filter ${\bf z}_k$ results in a different score function\footnote{The use of score function in this paper should not be confused with the scoring rule theory, which is used to design the learning objective of \textsc{Disco} Nets.} $\mathcal{G}_k({\bf y}; \boldsymbol{\theta}_{c}) \in \mathbb{R}^{B \times C}$. Note that we generate $K$ different score functions using $K$ different noise samples. These score functions are then used to sample corresponding bounding box labels $\hat{\bf y}_{c}^{k}$ such that all ground truth labels are present in it. This enables us to generate samples from the underlying distribution encoded by the network parameters. Note that obtaining a single sample is as efficient as a simple forward pass through the network. By placing the filters sufficiently far away from the output layer of the network, we can learn a highly non-linear mapping from the uniform distribution (used to generate the noise filter) to the output distribution (used to generate bounding box labels). 

\vspace{-2.5mm} 
\paragraph{Inference:} For the input pair $({\bf x}, {\bf z}_k)$, the classification branch of the conditional net outputs a score function $\mathcal{G}_k({\bf y}; \boldsymbol{\theta}_{c})$, which is a $B \times C$ matrix. The $(i,j)$-th element of the matrix, denoted by $\mathcal{G}_{k}^{(i,j)}$, denotes the score of the bounding box $i$ belonging to the category $j$. We will now redefine this score function such that it respects the constraints imposed by the annotation ${\bf a}$. In other words, for each category $j$ such that ${\bf a}^{(j)} =1$ there must exist at least one bounding box $i$ in ${\bf y}$ such that ${\bf y}^{(i)} = j$. The joint score for all the bounding box labels ${\bf y}$ is given by,
\begin{equation}
\label{eq2:redefiningScoreFn}
	S_k({\bf y}; \boldsymbol{\theta}_{c}) = \sum_{i=1}^B \mathcal{G}_{k}({\bf y}^{(i)}; \boldsymbol{\theta}_c) - H_k({\bf y}),
\end{equation}
where,
\begin{align}
\label{eq3:H_condition}
	H_k({\bf y}) = 
    		\begin{cases} 
            	0 
                	& \mathrm{if}\;\forall j \in \{1, \dots, C\}\ \mathrm{s.t.}\ {\bf a}^{(j)}=1,\\ 
                    &\; \exists i \in \{1, \dots, B\}\ \mathrm{s.t.}\ {\bf y}^{(i)}=j,	\\
                \infty &\; \mathrm{otherwise}.
            \end{cases}
\end{align}
Given the scoring function in equation (\ref{eq2:redefiningScoreFn}), we compute the $k$-th sample as
\begin{equation}
\label{eq4:samplingFromScoringFn}
	\hat{\bf y}_c^k = \argmax_{y \in \mathcal{Y}} S_k({\bf y}; \boldsymbol{\theta}_c).
\end{equation}
Note that in equation (\ref{eq4:samplingFromScoringFn}) the $\argmax$ needs to be computed over the entire output space $\mathcal{Y}$. A na\"ive brute force algorithm for this would be computationally infeasible. However, by using the structure of the higher order term $H_k$, we can design an efficient yet exact algorithm for equation (\ref{eq4:samplingFromScoringFn}).
Specifically,  we assign each bounding box proposal $i$ to its maximum scoring object class. If all the ground truth annotations ${\bf a}$ are not present in the generated bounding box labels, then we sample the bounding box which has the highest score corresponding to the foreground label, otherwise we sample all bounding boxes which satisfies the constraint. 

\vspace{-2.5mm} 
\section{Learning Objective} \label{sec4:learningObj}
In order to estimate the parameters of the prediction and conditional distribution, $\boldsymbol{\theta}_p$ and $\boldsymbol{\theta}_c$, we define a unified probabilistic learning objective based on the dissimilarity coefficient~\cite{rao1982diversity}. To this end, we require a task specific loss function, which we define next.

\vspace{-0.5mm} 
\subsection{Task Specific Loss Function} \label{ssec4.1:TaskLoss}
We define a loss function for object detection that decomposes over the bounding box proposals as follows:
\begin{equation}
\label{eq5:taskLoss}
	\Delta({\bf y}_1, {\bf y}_2) = \frac{1}{B} \sum_{i=1}^B \Delta({\bf y}_1^{(i)}, {\bf y}_2^{(i)}).
\end{equation}
Following the standard practice in most modern object detectors \cite{Huang_2017_CVPR}, $\Delta({\bf y}_1^{(i)}, {\bf y}_2^{(i)})$ is further decomposed as a weighted combination of the classification loss and the localization loss. We use $\lambda$ to denote the loss ratio ( ratio of the weight of localization loss to the weight of classification loss). We use a simple $0-1$ loss as our classification loss $\Delta_{cls}$, and $smoothL1$ \cite{girshick15fastrcnn} for our localization loss $\Delta_{loc}$. Formally, the task specific loss is given by,
\begin{equation}
\label{eq6:expandedTaskLoss}
	\Delta({\bf y}_1^{(i)}, {\bf y}_2^{(i)}) = \Delta_{cls}({\bf y}_1^{(i)}, {\bf y}_2^{(i)}) + \lambda  \Delta_{loc}({\bf y}_1^{(i)}, {\bf y}_2^{(i)}).
\end{equation}

\subsection{Objective Function} \label{ssec4.2:ObjFn}
The task of both the prediction distribution and the conditional distribution is to predict the bounding box labels. Moreover, as the conditional distribution utilizes the extra information in the form of the image-level label, it is expected to provide more accurate predictions for the bounding box labels ${\bf y}$. Leveraging on the task similarity between the two distributions, we would like to bring the two distributions close to each other, so that the extra knowledge of the conditional distribution can be transferred to the prediction distribution. Taking inspiration from~\cite{arun2018learning,kumar2012modeling}, we design a joint learning objective that can minimize the dissimilarity coefficient~\cite{rao1982diversity} between the prediction distribution and conditional distribution. In what follows, we briefly describe the concept of dissimilarity coefficient before applying it to our setting.

\vspace{-2.5mm} 
\paragraph{Dissimilarity Coefficient:} The dissimilarity coefficient between any two distributions $\Pr_1(\cdot)$ and $\Pr_2(\cdot)$ is determined by measuring their diversities. The diversity of a distribution $\Pr_{1}(\cdot)$ and a distribution $\Pr_{2}(\cdot)$ is defined as the expected difference between their samples, where the difference is measured by a task-specific loss function $\Delta'(\cdot, \cdot)$. Formally, we define the diversity as,
\begin{equation}
\label{eq7:DivCoeff}
    \begin{split}
	    DIV_{\Delta'}(\Pr\nolimits_1, \Pr\nolimits_2) =& \mathbb{E}_{{\bf y}_1 \sim \Pr_1(\cdot)} \big[ \mathbb{E}_{{\bf y}_2 \sim \Pr_2(\cdot)} \\
	     & \qquad\qquad\qquad [\Delta'({\bf y}_1, {\bf y}_2) ] \big].
	\end{split}
\end{equation}
If the model correctly brings the two distribution close to each other, we could expect the diversity $DIV_{\Delta'}(\Pr_1, \Pr_2)$ to be small. Using this definition of diversity, the dissimilarity coefficient of $\Pr_{1}$ and $\Pr_{2}$ is given by,
\begin{align}
\label{eq8:discoLoss}
	\begin{split}
		DISC_{\Delta'}(\Pr\nolimits_1, \Pr\nolimits_2) =& DIV_{\Delta'}(\Pr\nolimits_1, \Pr\nolimits_2) \\
		&- \gamma DIV_{\Delta'}(\Pr\nolimits_2, \Pr\nolimits_2) \\
        & - (1-\gamma)DIV_{\Delta'}(\Pr\nolimits_1, \Pr\nolimits_1),
	\end{split}
\end{align}
where $\gamma \in [0, 1]$. In other words, the dissimilarity coefficient between $\Pr_1$ and $\Pr_2$ is the difference between the diversity of $\Pr_1$ and $\Pr_2$, and a convex combination of their self-diversities. The self-diversity terms encourages the samples from each of the two distribution to be diverse, thus better representing the uncertainty of the task. In our experiments, we use $\gamma = 0.5$, which results in a symmetric dissimilarity coefficient between two distributions.

\vspace{-2.5mm}  
\paragraph{Learning Objective for Detection:} Given the above definition of dissimilarity coefficient, we can now specify our learning objective for the task specific loss $\Delta$ tuned for object detection (\ref{eq6:expandedTaskLoss}) as
\begin{equation}
\label{eq9:objectiveFn}
	\boldsymbol{\theta}_p^*, \boldsymbol{\theta}_c^* = \argmin_{\boldsymbol{\theta}_p, \boldsymbol{\theta}_c} DISC_{\Delta}(\Pr\nolimits_p(\boldsymbol{\theta}_p), \Pr\nolimits_c(\boldsymbol{\theta}_c)),
\end{equation}
where each of the diversity terms can be derived from equation (\ref{eq7:DivCoeff}). As discussed in Section~\ref{ssec3.2:probModel}, the conditional distribution is difficult to model directly. Therefore, the corresponding diversity terms are computed by stochastic estimators from $K$ samples $\hat{\bf y}_{c}^{k}$ of the conditional net. 
Thus, each of the diversity terms can be written as\footnote{Details in Appendix~\ref{ap:learning_objective}}
\begin{dmath}
	DIV_{\Delta}(\Pr\nolimits_p, \Pr\nolimits_c) =  \frac{1}{BK} \sum_{i=1}^B \sum_{k=1}^K \sum_{{\bf y}_p^{(i)}} \Pr\nolimits_p({\bf y}_p^{(i)}; \boldsymbol{\theta}_p)  \Delta({\bf y}_p^{(i)}, \hat{\bf y}_c^{k,(i)}), \label{eq10:crossDiv} 
\end{dmath}
\begin{dmath}
    DIV_{\Delta}(\Pr\nolimits_c, \Pr\nolimits_c) = \frac{1}{K(K-1)B} \sum_{\substack{{k,k'=1}\\{k' \neq k}}}^K \sum_{i=1}^B \Delta(\hat{\bf y}_c^{k,(i)}, \hat{\bf y'}_c^{k',(i)}), \label{eq11:selfCondDiv} 
\end{dmath}
\begin{dmath}
    DIV_{\Delta}(\Pr\nolimits_p, \Pr\nolimits_p) = \frac{1}{B} \sum_{i=1}^B \sum_{{\bf y}_p^{(i)}} \sum_{{\bf y'}_p^{(i)}} \Pr\nolimits_p({\bf y}_p^{(i)}; \boldsymbol{\theta}_p) \Pr\nolimits_p({\bf y'}_p^{(i)}; \boldsymbol{\theta}_p) \Delta({\bf y}_p^{(i)}, {\bf y'}_p^{(i)}). \label{eq12:selfPredDiv}
\end{dmath}
Here, $DIV_{\Delta}(\Pr\nolimits_p, \Pr\nolimits_c) $ measures the diversity between the prediction net and the conditional net, which is the expected difference between the samples from the two distributions as measured by the task specific loss function $\Delta$. Here $\Pr\nolimits_p$ is explicitly modeled, hence the expectation of its sample can be computed easily. However, as $\Pr\nolimits_c$ is not explicitly modeled, we compute the required expectation by drawing $K$ samples from the distribution. Likewise, $DIV_{\Delta}(\Pr\nolimits_c, \Pr\nolimits_c)$ measures the self diversity of the conditional net. We draw $K$ samples from the distribution to compute the required expectation. Also, the self diversity of the prediction net $DIV_{\Delta}(\Pr\nolimits_p, \Pr\nolimits_p)$ can be exactly computed as $\Pr\nolimits_p$ is explicitly modeled.

\section{Optimization} \label{sec5:Optimization}
As we employ deep neural networks to model the two distributions,  our objective function (\ref{eq9:objectiveFn}) is ideally suited to be minimized by stochastic gradient descent. While it may be possible to compute the gradients of both the networks simultaneously, in this work we use a simple coordinate descent optimization strategy. In more detail, the optimization proceeds by iteratively fixing the prediction network and learning the conditional network, followed by learning the prediction network for fixed conditional network.

The main advantage of using the iterative training strategy is that it results in an approach similar to the fully supervised learning of each network. This in turn allows us to readily use the algorithm developed in Fast-RCNN \cite{girshick15fastrcnn} and Discrete \textsc{Disco} Net \cite{bouchacourt2017thesis}. The outputs from the fixed network are treated as the pseudo ground truth bounding box labels for the other network. Furthermore, the iterative learning strategy also reduces the memory complexity of learning as only one network is trained at a time. 

\begin{figure*}[t]
\begin{center}
   \includegraphics[width=0.99\textwidth]{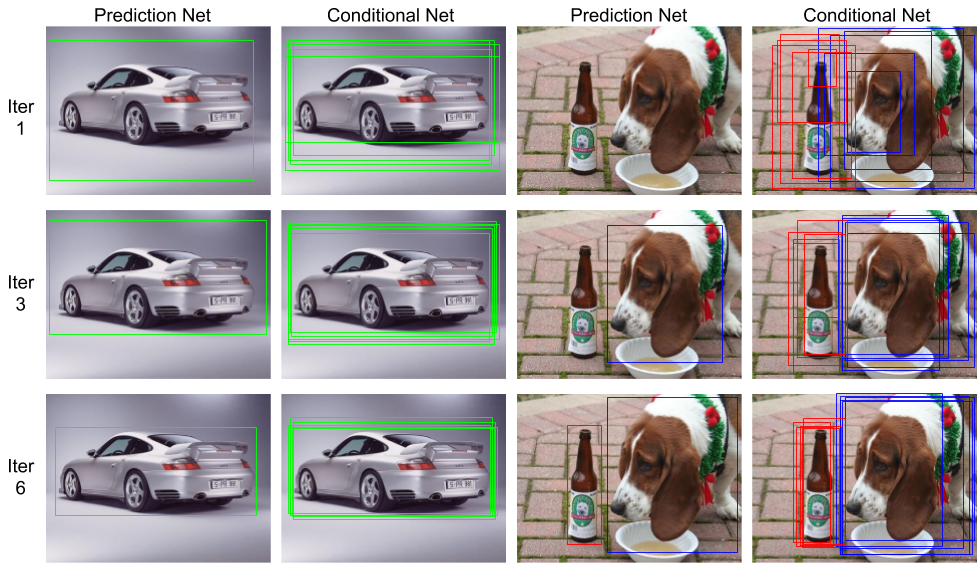}		
\end{center}
   \caption{\emph{Example of predictions of prediction net and conditional net. For prediction net, the visualization is after taking standard non maximal suppression using standard score threshold = $0.7$. Column 1 and 3 are output of the prediction network while column 2 and 4 are output from the conditional network. Row 1 represents prediction of the two networks after first iteration and row 2 and 3 represents prediction of the two networks after third and sixth (final) iteration respectively. Each object class is represented by different colored bounding box, where green box represents the \emph{car} category and red and blue represents the \emph{bottle} and \emph{dog} category respectively.}}
\label{fig2:VizOfLearningSteps}
\end{figure*}

Figure~\ref{fig2:VizOfLearningSteps} provides the visualization of the performance of the two networks over the different iterations of the iterative learning procedure. The estimated bounding box labels from the prediction net and those sampled from the conditional net for two images are depicted. For conditional net, we superimpose five different samples of bounding box labels. If all the samples agree with each other on bounding box labels, then the bounding boxes will have a high overlap, otherwise they will be scattered across the image. For visualization purposes only, a standard non maximal suppression (NMS) is applied with a score threshold of $0.7$ on the output of the prediction net. However, note that the non maximal suppression is not used during training of the prediction net. The two steps of the iterative algorithm are described below in brief. For completeness, the details are provided in appendix~\ref{ap:optimization}.


\subsection{Optimization over Prediction Distribution} \label{ssec5.1:OptPredNet}
For a fixed set of parameters $\boldsymbol{\theta}_c$ of the conditional network, the learning objective of the prediction net corresponds to the following:
\begin{equation}
\label{eq13:OptPredNet}
	\boldsymbol{\theta}_p^* = \argmin_{\boldsymbol{\theta}_p} DIV_{\Delta}(\Pr\nolimits_p, \Pr\nolimits_p) - (1-\gamma)DIV_{\Delta}(\Pr\nolimits_p, \Pr\nolimits_p).
\end{equation}
Note that, due to the use of dissimilarity coefficient, the above objective differs slightly from the one used for Fast-RCNN \cite{girshick15fastrcnn}. However, importantly, it is still differentiable with respect to $\boldsymbol{\theta}_p$. Hence, the prediction net can be directly optimized via stochastic gradient descent.

In order to visualize the optimization of the prediction net, let us consider Figure \ref{fig2:VizOfLearningSteps}. The first two columns show the bounding box labels from the prediction and the conditional nets for an image with single foreground object. As the image has a large foreground object with a clean background, both the prediction and the conditional nets have low uncertainty. This represents an easy case where the prediction net already has a high confidence for the bounding box labels in initial iterations, and therefore has little to gain from the conditional net. As expected, we see only a minor improvement in the predicted bounding box labels of the prediction net over the iterations.

The last two columns show bounding box labels from the prediction and conditional nets for a challenging image. The object \emph{dog} presents moderate difficulty to our algorithm, where initially the prediction net is highly uncertain while the conditional net has low uncertainty. After few iterations, the information present in the conditional net is successfully transferred over to the prediction net. This is shown in last row of the third column where the prediction net does a reasonable job at estimating the bounding boxes.

The second object \emph{bottle} in the image is a difficult example because of its small scale. We observe high uncertainty in both the networks. In such cases the prediction and the conditional nets will reject the bounding box labels having high diversity. Moreover, the uncertainty in the prediction net also decreases by learning from other easier instances of the object present in the data set.


\subsection{Optimization over Conditional Distribution} \label{ssec5.1:OptOverCondNet}
For a fixed set of parameters $\boldsymbol{\theta}_p$ of the prediction network, the learning objective for the conditional network corresponds to the following,
\begin{equation}
\label{eq14:OptCondNet}
	\boldsymbol{\theta}_c^* = \argmin_{\boldsymbol{\theta}_c} DIV_{\Delta}(\Pr\nolimits_p, \Pr\nolimits_c) - \gamma DIV_{\Delta}(\Pr\nolimits_c, \Pr\nolimits_c).
\end{equation}
The above objective function is similar to the one used in \cite{bouchacourt2017thesis} for supervised learning of Discrete \textsc{Disco} Nets. As our conditional net employs a sampling procedure over the scoring function $S_k({\bf y}; \boldsymbol{\theta}_c)$, objective~(\ref{eq14:OptCondNet}) is non-differentiable. However, as observed in \cite{bouchacourt2017thesis}, it is possible to compute an unbiased estimate of the gradients using the direct loss minimization technique \cite{hazan2010direct,song2016training}. Therefore, the conditional net can be optimized using stochastic gradient descent. We present the technical details of optimization, which are similar to those in \cite{bouchacourt2017thesis}, in appendix~\ref{ap:optimization}.

In order to visualize the optimization of the conditional net, let us first consider the easy case in Figure \ref{fig2:VizOfLearningSteps} (columns 1-2). Similar to the prediction net in the previous subsection, the uncertainty in the conditional net decreases marginally over the iterations, as it already has high confidence for the bounding box labels. For the challenging objects present in the image of the last two columns, we see that the prediction net has high uncertainty. The improvement in the predictions of the conditional net for these two cases are mainly attributed to the information gained by training on other easier examples of the \emph{dog} and the \emph{bottle} category present in the data set.

\section{Experiments} \label{sec5:experiments}
\subsection{Data set and Evaluation Metrics}
\paragraph{Data set:} We evaluate our method on the challenging PASCAL VOC 2007 and 2012 data sets~\cite{everingham2010pascal} which have $9,962$ and $22,531$ images respectively for $20$ object categories. These two data sets are divided into the train, val and test sets. Here we choose trainval set of $5011$ images for VOC 2007 and $11,540$ images for VOC 2012 to train our network. The trainval set is further split into $80\%-20\%$ to create new training and validation sets. We use a non-standard training-validation split in order to maximize the number of training images for our networks, while not overfitting our hyper-parameters on the test set. As we focus on weakly supervised detection, only image-level labels are utilized during training. 

\vspace{-2.5mm} 
\paragraph{Evaluation Metric} We use two metrics to evaluate our detection performance. First we evaluate detection using mean  Average Precision (mAP) on the PASCAL VOC 2007 and 2012 test sets, following the standard PASCAL VOC protocol~\cite{everingham2010pascal}. Second, we compute CorLoc~\cite{deselaers2012weakly} on the PASCAL VOC 2007 and 2012 trainval splits. CorLoc is the fraction of positive training images in which we localize an object of the target category correctly. Following~\cite{everingham2010pascal}, a detected bounding box is considered correct if it has at least 0.5 IoU with a ground truth bounding box.

\subsection{Implementation Details} 
We use standard Fast-RCNN~\cite{girshick15fastrcnn} to model prediction distribution and a modified Fast-RCNN to model the conditional distribution, as shown in Figure \ref{fig1:Arch}(a). We use the ImageNet pre-trained VGG16 Network~\cite{Simonyan14c} as the base CNN architecture for both our prediction and conditional nets.

The Fast-RCNN architecture is modified by adding a noise filter in its $5^{\mathrm{th}}$ conv-layer as an extra channel as shown in Figure~\ref{fig1:Arch}(b). A $1 \times 1$ filter is used to bring the number of channels back to the original dimensions ($512$ channels). No architectural changes are made for the prediction net. The bounding box proposals required for the Fast-RCNN is obtained from the Selective Search algorithm~\cite{uijlings2013selective}. Results based on the Region Proposal Networks are given in appendix~\ref{ap:implementation_details}.

Following the standard practice followed in Fast-RCNN, we train and test our method on a single scale. We also construct an ensemble by taking the ImageNet pre-trained VGG11 and VGG13 along with VGG16 and report its results. For all our experiments we choose $K=5$ for the conditional net. That is, we sample $5$ bounding boxes corresponding to $5$ noise filters, which are themselves sampled from a uniform distribution. For all other hyper-parameters, we use the same configurations as described in \cite{girshick15fastrcnn}. 

\begin{table*}[t]
\centering
\resizebox{\textwidth}{!}{%
\begin{tabular}{|l|cccccccccccccccccccc|c|}
\hline
\multicolumn{1}{|c|}{Method} & \multicolumn{1}{c}{aero} & \multicolumn{1}{c}{bike} & \multicolumn{1}{c}{bird} & \multicolumn{1}{c}{boat} & \multicolumn{1}{c}{bottle} & \multicolumn{1}{c}{bus} & \multicolumn{1}{c}{car} & \multicolumn{1}{c}{cat} & \multicolumn{1}{c}{chair} & \multicolumn{1}{c}{cow} & \multicolumn{1}{c}{table} & \multicolumn{1}{c}{dog} & \multicolumn{1}{c}{horse} & \multicolumn{1}{c}{mbike} & \multicolumn{1}{c}{pson} & \multicolumn{1}{c}{plant} & \multicolumn{1}{c}{sheep} & \multicolumn{1}{c}{sofa} & \multicolumn{1}{c}{train} & \multicolumn{1}{c}{tv} & \multicolumn{1}{|c|}{mAP} \\ \hline
WSDDN~\cite{Bilen_2016_CVPR} & 46.4 & 58.3 & 35.5 & 25.9 & 14.0 & 66.7 & 53.0 & 39.2 & 8.9 & 41.8 & 26.6 & 38.6 & 44.7 & 59.0 & 10.8 & 17.3 & 40.7 & 49.6 & 56.9 & 50.8 & 39.3    \\
WSCCN~\cite{Diba_2017_CVPR} & 49.5 & 60.6 & 38.6 & 29.2 & 16.2 & 70.8 & 56.9 & 42.5 & 10.9 & 44.1 & 29.9 & 42.2 & 47.9 & 64.1 & 13.8 & 23.5 & 45.9 & 54.1 & 60.8 & 54.5 & 42.8 \\
k-EM~\cite{yan2017weakly} & 59.8 & 64.6 & 47.8 & 28.8 & 21.4 & 67.7 & 70.3 & 61.2 & 17.2 & 51.5 & 34.0 & 42.3 & 48.8 & 65.9 & 9.3  & 21.1 & 53.6 & 51.4 & 54.7 & 50.7 & 46.1 \\
OICR~\cite{Tang_2017_CVPR} & 65.5 & 67.2 & 47.2 & 21.6 & 22.1 & 68.0 & 68.5 & 35.9 & 5.7 & 63.1 & 49.5 & 30.3 & 64.7 & 66.1 & 13.0 & 25.6 & 50.0 & 57.1 & 60.2 & 59.0 & 47.0 \\
ZLDN~\cite{Zhang_2018_CVPR} & 55.4 & 68.5 & 50.1 & 16.8 & 20.8 & 62.7 & 66.8 & 56.5 & 2.1 & 57.8 & 47.5 & 40.1 & 69.7 & 68.2 & 21.6 & 27.2 & 53.4 & 56.1 & 52.5 & 58.2 & 47.6 \\
CL~\cite{wang2018collaborative} & 61.2 & 66.6 & 48.3 & 26.0 & 15.8 & 66.5 & 65.4 & 53.9 & 24.7 & 61.2 & 46.2 & 53.5 & 48.5 & 66.1 & 12.1 & 22.0 & 49.2 & 53.2 & 66.2 & 59.4 & 48.3 \\
ML-LocNet~\cite{Zhang_2018_ECCV} & 60.8 & 70.6 & 47.8 & 30.2 & 24.8 & 64.9 & 68.4 & 57.9 & 11.0 & 51.3 & 55.5 & 48.1 & 68.7 & 69.5 & 28.3 & 25.2 & 51.3 & 56.5 & 60.0 & 43.1 & 49.7 \\
WS-RPN~\cite{Tang_2018_ECCV} & 63.0 & 69.7 & 40.8 & 11.6 & 27.7 & 70.5 & 74.1 & 58.5 & 10.0 & 66.7 & 60.6 & 34.7 & 75.7 & 70.3 & 25.7 & 26.5 & 55.4 & 56.4 & 55.5 & 54.9 & 50.4 \\
W2F~\cite{Zhang_2018_CVPR_W2F} & 63.5 & 70.1 & 50.5 & 31.9 & 14.4 & 72.0 & 67.8 & 73.7 & 23.3 & 53.4 & 49.4 & 65.9 & 57.2 & 67.2 & 27.6 & 23.8 & 51.8 & 58.7 & 64.0 & 62.3 & 52.4 \\
\hline
Pred Net (VGG) & 66.7 & 69.5 & 52.8 & 31.4 & 24.7 & 74.5 & 74.1 & 67.3 & 14.6 & 53.0 & 46.1 & 52.9 & 69.9 & 70.8 & 18.5 & 28.4 & 54.6 & 60.7 & 67.1 & 60.4 & \textbf{52.9}  \\
Pred Net (Ens) & 67.7 & 70.4 & 52.9 & 31.3 & 26.1 & 75.5 & 73.7 & 68.6 & 14.9 & 54.0 & 47.3 & 53.7 & 70.8 & 70.2 & 19.7 & 29.2 & 54.9 & 61.3 & 67.6 & 61.2 & \textbf{53.6}\\
\hline
\end{tabular}
}
\caption{Detection average precision ($\%$) for different methods on VOC 2007 test set.}
\label{table1:mAP07}
\end{table*}

\begin{table*}[t]
\centering
\resizebox{\textwidth}{!}{%
\begin{tabular}{|l|cccccccccccccccccccc|c|}
\hline
\multicolumn{1}{|c|}{Method} & \multicolumn{1}{c}{aero} & \multicolumn{1}{c}{bike} & \multicolumn{1}{c}{bird} & \multicolumn{1}{c}{boat} & \multicolumn{1}{c}{bottle} & \multicolumn{1}{c}{bus} & \multicolumn{1}{c}{car} & \multicolumn{1}{c}{cat} & \multicolumn{1}{c}{chair} & \multicolumn{1}{c}{cow} & \multicolumn{1}{c}{table} & \multicolumn{1}{c}{dog} & \multicolumn{1}{c}{horse} & \multicolumn{1}{c}{mbike} & \multicolumn{1}{c}{pson} & \multicolumn{1}{c}{plant} & \multicolumn{1}{c}{sheep} & \multicolumn{1}{c}{sofa} & \multicolumn{1}{c}{train} & \multicolumn{1}{c}{tv} & \multicolumn{1}{|c|}{mean} \\ \hline
WSCCN~\cite{Diba_2017_CVPR} &  83.9 & 72.8 & 64.5 & 44.1 & 40.1 & 65.7 & 82.5 & 58.9 & 33.7 & 72.5 & 25.6 & 53.7 & 67.4 & 77.4 & 26.8 & 49.1 & 68.1 & 27.9 & 64.5 & 55.7 & 56.7 \\
WSDDN~\cite{Bilen_2016_CVPR} & 68.9 & 68.7 & 65.2 & 42.5 & 40.6 & 72.6 & 75.2 & 53.7 & 29.7 & 68.1 & 33.5 & 45.6 & 65.9 & 86.1 & 27.5 & 44.9 & 76.0 & 62.4 & 66.3 & 66.8 & 58.0   \\
ZLDN~\cite{Zhang_2018_CVPR} & 74.0 & 77.8 & 65.2 & 37.0 & 46.7 & 75.8 & 83.7 & 58.8 & 17.5 & 73.1 & 49.0 & 51.3 & 76.7 & 87.4 & 30.6 & 47.8 & 75.0 & 62.5 & 64.8 & 68.8 & 61.2 \\
OICR~\cite{Tang_2017_CVPR} & 85.8 & 82.7 & 62.8 & 45.2 & 43.5 & 84.8 & 87.0 & 46.8 & 15.7 & 82.2 & 51.0 & 45.6 & 83.7 & 91.2 & 22.2 & 59.7 & 75.3 & 65.1 & 76.8 & 78.1 & 64.3 \\
CL~\cite{wang2018collaborative} & 85.8 & 80.4 & 73.0 & 42.6 & 36.6 & 79.7 & 82.8 & 66.0 & 34.1 & 78.1 & 36.9 & 68.6 & 72.4 & 91.6 & 22.2 & 51.3 & 79.4 & 63.7 & 74.5 & 74.6 & 64.7 \\
k-EM~\cite{yan2017weakly} & 79.8 & 77.8 & 66.7 & 50.3 & 57.0 & 80.1 & 89.9 & 71.5 & 29.9 & 75.9 & 30.5 & 58.9 & 73.2 & 90.2 & 25.4 & 51.8 & 80.2 & 60.3 & 72.4 & 78.9 & 65.0 \\
WS-RPN~\cite{Tang_2018_ECCV} & 83.8 & 82.7 & 60.7 & 35.1 & 53.8 & 82.7 & 88.6 & 67.4 & 22.0 & 86.3 & 68.8 & 50.9 & 90.8 & 93.6 & 44.0 & 61.2 & 82.5 & 65.9 & 71.1 & 76.7 & 68.4 \\
ML-LocNet~\cite{Zhang_2018_ECCV} &  81.7 & 82.9 & 68.7 & 44.4 & 53.9 & 80.3 & 88.9 & 70.5 & 32.6 & 74.0 & 62.7 & 61.7 & 81.4 & 91.6 & 46.0 & 60.6 & 75.2 & 69.2 & 78.7 & 65.8 & 68.6 \\
W2F~\cite{Zhang_2018_CVPR_W2F} & 85.4 & 87.5 & 62.5 & 54.3 & 35.5 & 85.3 & 86.6 & 82.3 & 39.7 & 82.9 & 49.4 & 76.5 & 74.8 & 90.0 & 46.8 & 53.9 & 84.5 & 68.3 & 79.1 & 79.9 & 70.3 \\
\hline
Pred Net (VGG) & 88.6 & 86.3 & 71.8 & 53.4 & 51.2 & 87.6 & 89.0 & 65.3 & 33.2 & 86.6 & 58.8 & 65.9 & 87.7 & 93.3 & 30.9 & 58.9 & 83.4 & 67.8 & 78.7 & 80.2 & \textbf{70.9} \\
Pred Net (Ens) & 89.2 & 86.7 & 72.2 & 50.9 & 51.8 & 88.3 & 89.5 & 65.6 & 33.6 & 87.4 & 59.7 & 66.4 & 88.5 & 94.6 & 30.4 & 60.2 & 83.8 & 68.9 & 78.9 & 81.3 & \textbf{71.4} \\
\hline
\end{tabular}
}
\caption{CorLoc (in \%) for different methods on VOC 2007 trainval set.}
\label{table2:corLoc07}
\end{table*}


\subsection{Results}
In this subsection, we will first compare our method with existing state-of-the-art methods for detection and correct localization tasks on VOC 2007 and 2012 data sets. Then through ablation experiments, see how various terms of our dissimilarity coefficient based objective function contribute towards the accuracy gained. We present further ablation studies in appendix~\ref{ap:experiments}.

\begin{table}[t]
\centering
\resizebox{\columnwidth}{!}{%
\begin{tabular}{|l|c|c|c|c|c|c|}
\hline
Method & WSCCN~\cite{Diba_2017_CVPR} & DSL~\cite{Jie_2017_CVPR} & OICR~\cite{Tang_2017_CVPR} & W2F~\cite{Zhang_2018_CVPR_W2F} & PredNet(VGG) & PredNet(Ens) \\
\hline
mAP $\%$ & 37.9 & 38.3 & 42.5 & 47.8 & \textbf{48.4} & \textbf{49.5} \\
CorLoc $\%$ & - & 58.8 & 65.6 & 69.4 & \textbf{69.5} & \textbf{70.2} \\
\hline
\end{tabular}
}
\caption{Results for different methods on VOC 2012. See appendix~\ref{ap:experiments} for details.}
\label{table3:VOC12}
\vspace{-5mm}  
\end{table}

\vspace{-2.5mm} 
\subsubsection{Comparison with other methods} 
We compare our proposed method with other state-of-the-art weakly supervised methods. The detection average precision (AP) and correct localization (CorLoc) on the PASCAL VOC 2007 and 2012 data sets are shown in Table~\ref{table1:mAP07}, Table~\ref{table2:corLoc07} and Table~\ref{table3:VOC12} respectively. Compared with the other methods, our proposed framework achieves state-of-the-art performance using a single model. 

Compared to the state-of-the-art method, if we were to only train and test Zhang \emph{et al.}~\cite{Zhang_2018_CVPR_W2F} (W2F) using a single scale, where they achieve 49.0\% mAP, we get an improvement of 3.9\%. Our framework trained on a single scale still outperforms W2F by 0.5\% even when they train and test using multiple scales. We approximate the use of multiple scales by ensembling, which gives us a further improvement over the state-of-the-art method by over 1.2\%. 

The weakly supervised detector employed in W2F models the annotation constraint using a fully factorized distribution. We argue that our choice of modeling the annotation aware conditional distribution exactly but efficiently, using Discrete \textsc{Disco} Net, gives us the improved performance. Moreover, unlike W2F, our method combines the weakly supervised and the strongly supervised detectors with a single learning objective instead of training them in a non-end-to-end, cascaded fashion. We note that the pseudo-ground-truth excavation (PGE) algorithm proposed in W2F is complementary to our method, and can also be employed over the samples generated from conditional distribution to further improve the accuracy of our method.

\vspace{-3mm} 
\subsubsection{Effect of the diversity coefficient terms} 
In order to understand the effect of various diversity coefficient terms in our objective (\ref{eq8:discoLoss}), we remove the self-diversity term in one or both of our probabilistic networks ($\Pr_c$ and $\Pr_p$). In order to obtain a single sample from our conditional network, we feed a zero noise vector (denoted by $PW_c$). The prediction network still outputs the probability of each bounding box belonging to each class. However, by removing the self-diversity term, we encourage it to output a peakier distribution (denoted by $PW_p$). Table~\ref{table4:ablation} shows that both the self-diversity terms are important to obtain the maximum accuracy. Relatively speaking, it is more important to include the self-diversity in the conditional network in order to deal with the difficult examples (example, bottle in figure~\ref{fig2:VizOfLearningSteps}). Moreover, this enforces a diverse set of outputs from the conditional network, which helps the prediction network to avoid overfitting the samples during training.

\begin{table}[t]
\centering
\resizebox{\columnwidth}{!}{%
\begin{tabular}{|c|c|c|c|c|}
\hline
Method         & \begin{tabular}[c]{@{}c@{}}$\Pr\nolimits_{p}, \Pr\nolimits_{c}$\\ (proposed)\end{tabular} & $\Pr\nolimits_{p}, PW_c$ & $PW_p, \Pr\nolimits_{c}$ & $PW_p, PW_c$ \\ \hline
Mean AP & 52.9 & 50.1 & 52.6  & 49.5 \\ \hline
\end{tabular}
}
\caption{Detection Average Precision ($\%$) for various ablative settings on VOC 2007 test set}
\label{table4:ablation}
\vspace{-5mm}
\end{table}

\vspace{-2.5mm} 
\section{Discussion}
We presented a novel framework to train an object detector using a weakly supervised data set. Our framework employs a probabilistic objective based on dissimilarity coefficient to model the uncertainty in the location of objects. We show that explicitly modeling the complex non-factorizable conditional distribution is a necessary modeling choice and present an efficient mechanism based on a discrete generative model, the Discrete \textsc{Disco} Nets, to do so. Extensive experiments on the benchmark data sets have shown that our framework successfully transfers the information present in the image-level annotations for the task of object detection.

In future, we would like to investigate the use of active learning, to further benefit our network in terms of the accuracy of the fully supervised annotations. This will help bridge the performance gap between the strongly supervised detectors and detectors trained using low-cost annotations.



\clearpage

{\small
\bibliographystyle{ieee}
\bibliography{egbib}
}

\clearpage

\begin{appendices}

\section{Learning Objective}
\label{ap:learning_objective}
In this section we provide detailed derivation of the objective function presented in Section~\ref{ssec4.2:ObjFn} of the paper.

Given the loss function $\Delta$ (equation (\ref{eq6:expandedTaskLoss}) of main paper), which is tuned for the task of object detection, we compute the diversity terms as given in equation (\ref{eq7:DivCoeff}) of the main paper. Recall that the diversity for any two distributions is the expected loss of the samples drawn from the two distributions. For the prediction distribution $\Pr\nolimits_p$ and the conditional distribution $\Pr\nolimits_c$, we derive the diversity between them and their self diversities as follows.

\paragraph{Diversity between prediction net and conditional net:} Following equation (\ref{eq7:DivCoeff}) of the main paper, the diversity between prediction and conditional distribution can be written as, 
\begin{dmath}
	DIV_{\Delta}(\Pr\nolimits_p, \Pr\nolimits_c) = \mathbb{E}_{{\bf y}_p \sim \Pr_p({\bf y} | {\bf x}; \boldsymbol{\theta}_p)} \big[ \mathbb{E}_{{\bf y}_c \sim \Pr_c({\bf y} | {\bf x, h}; \boldsymbol{\theta}_c)} [ \Delta({\bf y}_p, {\bf y}_c) ] \big].
\end{dmath}
The task specific loss function is decomposed over the bounding boxes as given in equation (\ref{eq5:taskLoss}) of the main paper. We then write the expectation with respect to the conditional distribution (the inner distribution) as expectation over the random variables ${\bf z}$ with distribution $\Pr({\bf z})$ using Law of the Unconscious Statistician (LOTUS). 
\begin{dmath}
	DIV_{\Delta}(\Pr\nolimits_p, \Pr\nolimits_c) = \mathbb{E}_{{\bf y}_p \sim \Pr\nolimits_p({\bf y} | {\bf x}; \boldsymbol{\theta}_p)} \big[ \mathbb{E}_{{\bf z} \sim \Pr({\bf z})} [ \frac{1}{B} \sum_{i=1}^B \Delta({\bf y}_p^{(i)}, \hat{\bf y}_c^{k,(i)}) ] \big].
\end{dmath}
The expectation over the random variable ${\bf z}$ with distribution $\Pr({\bf z})$ is approximated by taking $K$ samples from $\Pr({\bf z})$,
\begin{dmath}
	DIV_{\Delta}(\Pr\nolimits_p, \Pr\nolimits_c) = \mathbb{E}_{{\bf y}_p \sim \Pr\nolimits_p({\bf y} | {\bf x}; \boldsymbol{\theta}_p)} \big[ \frac{1}{K} \sum_{k=1}^K \frac{1}{B} \sum_{i=1}^B \Delta({\bf y}_p^{(i)}, \hat{\bf y}_c^{k,(i)}) \big].
\end{dmath}
We finally compute the expectation with respect to the prediction distribution as,
\begin{dmath}
	DIV_{\Delta}(\Pr\nolimits_p, \Pr\nolimits_c) = \frac{1}{BK} \sum_{i=1}^B  \sum_{k=1}^K \sum_{{\bf y}_p^{(i)}} \Pr\nolimits_p({\bf y}_p^{(i)}; \boldsymbol{\theta}_p)  \Delta({\bf y}_p^{(i)}, \hat{\bf y}_c^{k,(i)}).
\end{dmath}

\paragraph{Self diversity for conditional net:} As above, using equation (\ref{eq7:DivCoeff}) of the main paper, we write the self diversity coefficient of the conditional distribution as
\begin{dmath}
	DIV_{\Delta}(\Pr\nolimits_c, \Pr\nolimits_c)  = \mathbb{E}_{{\bf y}_c \sim \Pr\nolimits_c({\bf y} | {\bf x, h}; \boldsymbol{\theta}_c)} \big[ \mathbb{E}_{{\bf y}_c' \sim \Pr\nolimits_c({\bf y} | {\bf x, h}; \boldsymbol{\theta}_c)} [ \Delta({\bf y}_c, {\bf y}_c') ] \big].
\end{dmath}
We now write the two expectations with respect to the conditional distribution as the expectation over the random variables ${\bf z}$ and ${\bf z}'$ respectively. The task specific loss function is decomposed over the bounding box as shown in equation (\ref{eq5:taskLoss}) of the main paper. Therefore, we re-write the above equation as
\begin{dmath}
	DIV_{\Delta}(\Pr\nolimits_c, \Pr\nolimits_c) = \mathbb{E}_{{\bf z} \sim \Pr({\bf z})} \big[ \mathbb{E}_{{\bf z}' \sim \Pr({\bf z})} [ \frac{1}{B} \sum_{i=1}^B \Delta(\hat{\bf y}_c^{k,(i)}, \hat{\bf y'}_c^{k,(i)}) ] \big].
\end{dmath}
In order to approximate the expectation over the random variables ${\bf z}$ and ${\bf z}'$, we use $K$ samples from the distribution $\Pr({\bf z})$ as
\begin{dmath}
	DIV_{\Delta}(\Pr\nolimits_c, \Pr\nolimits_c) = \frac{1}{K} \sum_{k=1}^K \frac{1}{K-1} \sum_{\substack{ k'=1, \\ k' \neq k}}^K \frac{1}{B} \sum_{i=1}^B \Delta(\hat{\bf y}_c^{k,(i)}, \hat{\bf y'}_c^{k',(i)}).
\end{dmath}
On re-arranging the above equation, we get
\begin{dmath}
	DIV_{\Delta}(\Pr\nolimits_c, \Pr\nolimits_c) = \frac{1}{K(K-1)B} \sum_{\substack{k,k'=1^K \\ k' \neq k}}^K \sum_{i=1}^B \Delta(\hat{\bf y}_c^{k,(i)}, \hat{\bf y'}_c^{k',(i)}).
\end{dmath}

\paragraph{Self diversity for prediction net:} Similar to the above two cases, using equation (\ref{eq7:DivCoeff}) of the main paper, we can write the self diversity of the prediction net as
\begin{dmath}
	DIV_{\Delta}(\Pr\nolimits_p, \Pr\nolimits_p) = \mathbb{E}_{{\bf y}_p \sim \Pr\nolimits_p({\bf y} | {\bf x}; \boldsymbol{\theta}_p)} \big[ \mathbb{E}_{{\bf y'}_p \sim \Pr\nolimits_p({\bf y} | {\bf x}; \boldsymbol{\theta}_p)} [ \Delta({\bf y}_p, {\bf y'}_p) ] \big].
\end{dmath}
We then decompose the task specific loss function over the bounding boxes as described in equation (\ref{eq5:taskLoss}) of the main paper,
\begin{dmath}
	DIV_{\Delta}(\Pr\nolimits_p, \Pr\nolimits_p) = \mathbb{E}_{{\bf y}_p \sim \Pr\nolimits_p({\bf y} | {\bf x}; \boldsymbol{\theta}_p)} \big[ \mathbb{E}_{{\bf y'}_1 \sim \Pr\nolimits_p({\bf y} | {\bf x}; \boldsymbol{\theta}_p)} [ \frac{1}{B} \sum_{i=1}^B \Delta({\bf y}_p^{(i)}, {\bf y'}_p^{(i)}) ] \big]
\end{dmath}

Note that the prediction distribution is a fully factorized distribution, and we can compute its exact expectation. Therefore, we compute the two expectations with respect to the prediction distribution as,
\begin{align}
     &\mathbb{E}_{{\bf y}_p \sim \Pr\nolimits_p({\bf y'} | {\bf x}; \boldsymbol{\theta}_p)} \big[ \frac{1}{B} \sum_{i=1}^B \sum_{{\bf y'}_p^{(i)}} \Pr\nolimits_p({\bf y'}_p^{(i)}; \boldsymbol{\theta}_p) \Delta({\bf y}_p^{(i)}, {\bf y'}_p^{(i)}) \big] \nonumber \\
    &= \frac{1}{B} \sum_{i=1}^B \sum_{{\bf y}_p^{(i)}} \sum_{{\bf y'}_p^{(i)}} \Pr\nolimits_p({\bf y}_p^{(i)}; \boldsymbol{\theta}_1) \Pr\nolimits_p({\bf y'}_p^{(i)}; \boldsymbol{\theta}_p) \Delta({\bf y}_p^{(i)}, {\bf y'}_p^{(i)})
\end{align}

\section{Optimization}
\label{ap:optimization}
\subsection{Optimization over Prediction Distribution}
As parameters $\boldsymbol{\theta}_c$ of the conditional distribution are constant, the learning objective of the prediction distribution (equation (\ref{eq13:OptPredNet}) of the main paper) results in a fully supervised training of the Fast-RCNN network~\cite{girshick15fastrcnn}. Note that the only difference between training of a standard Fast-RCNN architecture and our prediction net is the use of the dissimilarity objective function (equation (\ref{eq13:OptPredNet}) of the main paper) instead of minimizing the multi-task loss of the Fast-RCNN.

The prediction net takes as the input an image and the $K$ predictions sampled from the conditional net. Treating these predictions of the conditional net as the pseudo ground truth label, we compute the gradient of our dissimilarity coefficient based loss function. As the objective given in equation (\ref{eq13:OptPredNet}) of the main paper is differentiable with respect to parameters $\boldsymbol{\theta}_p$, we update the network by employing stochastic gradient descent.

\subsection{Optimization over Conditional Distribution}

\paragraph{A non-differentiable training procedure:} The conditional net is modeled using a Discrete \textsc{Disco} Net which employs a sampling step from the scoring function $S_k({\bf y}; \boldsymbol{\theta}_{c})$. This sampling step makes the objective function non-differentiable with respect to the parameters $\boldsymbol{\theta}_c$, even though the scoring function $S_k({\bf y}; \boldsymbol{\theta}_{c})$ itself is differentiable. However, as the prediction network is fixed, the above objective function reduces to the one used in Bouchacourt \emph{et al.} \cite{bouchacourt2017thesis} for fully supervised training. Therefore, similar to Bouchacourt \emph{et al.}~\cite{bouchacourt2017thesis} we solve this problem by estimating the gradients of our objective function with the help of temperature parameter $\epsilon$ as,
\begin{dmath}
		\nabla_{\boldsymbol{\theta}_c} DISC_{\Delta}^{\epsilon}(\Pr\nolimits_p(\boldsymbol{\theta}_p), \Pr\nolimits_c(\boldsymbol{\theta}_c)) =  \pm \lim_{\epsilon \rightarrow 0} \frac{1}{\epsilon}\big( DIV_{\Delta}^{\epsilon}(\Pr\nolimits_p, \Pr\nolimits_c) 
        - \gamma DIV_{\Delta}^{\epsilon}(\Pr\nolimits_c, \Pr\nolimits_c)\big)
\label{eq26:gradCondNet}
\end{dmath}
where,
\begin{dmath}
		DIV_{\Delta}^{\epsilon}(\Pr\nolimits_p, \Pr\nolimits_c) = \mathbb{E}_{{\bf y}_p \sim \Pr\nolimits_p(\boldsymbol{\theta}_p)} \big[ \mathbb{E}_{{\bf z}_k \sim \Pr({\bf z})} [ \nabla_{\boldsymbol{\theta}_c} S_k(\hat{\bf y}_a; \boldsymbol{\theta}_c) - \nabla_{\boldsymbol{\theta}_c} S_k(\hat{\bf y}_c; \boldsymbol{\theta}_c) ] \big] 
\end{dmath}
\begin{dmath}
        DIV_{\Delta}^{\epsilon}(\Pr\nolimits_c, \Pr\nolimits_c) =  \mathbb{E}_{{\bf z}_k \sim \Pr({\bf z})} \big[ \mathbb{E}_{{{\bf z}'}_k \sim \Pr({\bf z})} [ \nabla_{\boldsymbol{\theta}_c} S_k(\hat{\bf y}_b; \boldsymbol{\theta}_c) - \nabla_{\boldsymbol{\theta}_c} S_k({\hat{\bf y}'}_c; \boldsymbol{\theta}_c) ] \big]
\label{eq28:expGradCondNet}
\end{dmath}
and,
\begin{equation}
\label{eq29:sampleGradCondNet}
	\begin{split}
		\hat{\bf y}_c & = \argmax_{y \in \mathcal{Y}} S_k({\bf y}; \boldsymbol{\theta}_c) \\
        {\hat{\bf y}'}_c & = \argmax_{y \in \mathcal{Y}} S_{k'}({\bf y}; \boldsymbol{\theta}_c) \\
        \hat{\bf y}_a & = \argmax_{y \in \mathcal{Y}} S_k({\bf y}; \boldsymbol{\theta}_c) \pm \epsilon \Delta({\bf y}_p, \hat{\bf y}_c) \\
        \hat{\bf y}_b & = \argmax_{y \in \mathcal{Y}} S_k({\bf y}; \boldsymbol{\theta}_c) \pm \epsilon \Delta(\hat{\bf y}_c, {\hat{\bf y}'}_c)
	\end{split}
\end{equation}
In our experiments, we fix the temperature parameter $\epsilon$ as, $\epsilon = +1$.

\paragraph{Intuition for the gradient computation:} We now present an intuitive explanation of the computation of gradient, as given in equation (\ref{eq26:gradCondNet}). For an input ${\bf x}$ and two noise samples ${\bf z}_k, {\bf z}_{k'}$, the conditional net outputs two scores $ S_{k}({\bf y}; \boldsymbol{\theta}_c)$ and $ S_{k'}({\bf y}; \boldsymbol{\theta}_c)$, with the corresponding maximum scoring outputs $\hat{\bf y}_c$ and ${\hat{\bf y}'}_c$. The model parameters $\boldsymbol{\theta}_c$ are updated via gradient descent in the negative direction of $\nabla_{\boldsymbol{\theta}_c} DISC_{\Delta}^{\epsilon}(\Pr\nolimits_p(\boldsymbol{\theta}_p), \Pr\nolimits_c(\boldsymbol{\theta}_c))$.
\begin{itemize}
\item The term $DIV_{\Delta}^{\epsilon}(\Pr\nolimits_p, \Pr\nolimits_c)$ updates the model parameters towards the maximum scoring prediction $\hat{\bf y}_c$ of the score $ S_{k}({\bf y}; \boldsymbol{\theta}_c)$ while moving away from $\hat{\bf y}_a$, where $\hat{\bf y}_a$ is the sample corresponding to the maximum loss augmented score $S_{k}({\bf y}; \boldsymbol{\theta}_c) \pm \epsilon \Delta({\bf y}_p, \hat{\bf y}_c)$ with respect to the fixed prediction distribution samples ${\bf y}_p$. This encourages the model to move away from the prediction providing high loss with respect to the pseudo ground truth labels.
\item The term $\gamma DIV_{\Delta}^{\epsilon}(\Pr\nolimits_c, \Pr\nolimits_c)$ updates the model towards ${\bf y}_b$ and away from the $\hat{\bf y}_c$. Note the two negative signs giving the update in the positive direction. Here ${\bf y}_b$ is the sample corresponding to the maximum loss augmented score $S_k({\bf y}; \boldsymbol{\theta}_c) \pm \epsilon \Delta(\hat{\bf y}_c, \hat{\bf y}')$ with respect to the other prediction ${\hat{\bf y}'}_c$, encouraging diversity between $\hat{\bf y}_c$ and ${\hat{\bf y}'}_c$.
\end{itemize}


\paragraph{Training algorithm for conditional net:} Pseudo-code for training the conditional network for a single sample from weakly supervised data is presented in algorithm~\ref{algo1:condNet} below. In algorithm~\ref{algo1:condNet}, statements 1 to 3 describe the sampling process and computing the loss augmented prediction. We first sample $K$ different predictions $\hat{\bf y}_c^k$ corresponding to each noise vector ${\bf z}_k$ in statement 2. For the sampled prediction $\hat{\bf y}_c^k$ we compute the maximum loss augmented score $S_{k}({\bf y}; \boldsymbol{\theta}_c) \pm \epsilon \Delta({\bf y}_p, \hat{\bf y}_c)$. This is then used to find the loss augmented prediction $\hat{\bf y}_a$ given in statement 3. 

In order to compute the gradients of the self diversity of conditional distribution, we need to find the maximum loss augmented prediction ${\bf y}_b$. Here, the loss is computed between a pair of $K$ different predictions of the conditional net that we have already obtained. This is shown by statements 4 to 7 in algorithm~\ref{algo1:condNet}.

For the purpose of optimizing the conditional net using gradient descent, we need to find the gradients for the objective function of the conditional net defined in equation (\ref{eq14:OptCondNet}) of the main paper. The computation of the unbiased approximate gradients for the individual terms in the objective function is shown in statement 8. We finally optimize the conditional net by the employing gradient descent step and updating the model parameters by descending to the approximated gradients as shown in statement 9 of algorithm~\ref{algo1:condNet}.

\begin{algorithm}[h]
\RestyleAlgo{boxed}
\DontPrintSemicolon
\SetAlgoLined
\SetKwInOut{Input}{Input}\SetKwInOut{Output}{Output}

\Input{Training input $({\bf x}, {\bf a}) \in \mathcal{W}$, and prediction net output ${\bf y}_p$}
\Output{$\hat{\bf y}_c^{1}, \dots, \hat{\bf y}_c^K$, sample $K$ predictions from the model}
\BlankLine
\For{$k = 1 \dots K$}
{
	Sample noise vector ${\bf z}_k$, generate output $\hat{\bf y}_c^k$:
    \[ \hat{\bf y}_c^k = \argmax_{y \in \mathcal{Y}} S_k({\bf y}; \boldsymbol{\theta}_c) \] 
    
    Find loss augmented prediction $\hat{\bf y}_a^k$ w.r.t. output from prediction net ${\bf y}_p$:
    \[ \hat{\bf y}_a^k = \argmax_{y \in \mathcal{Y}} S_k({\bf y}; \boldsymbol{\theta}_c) \pm \epsilon \Delta({\bf y}_p, \hat{\bf y}_c^k) \] 

}

Compute loss augmented predictions:\\
\For{$k = 1,\dots,K$}
{
	\For{$k' = 1,\dots,K, k' \neq k$}
    {
    	Find loss augmented prediction $\hat{\bf y}_b^k$ w.r.t. other conditional net outputs $\hat{\bf y}_c^k$:
        \[ \hat{\bf y}_b^{k,k'} = \argmax_{y \in \mathcal{Y}} S_k({\bf y}; \boldsymbol{\theta}_c) \pm \epsilon \Delta(\hat{\bf y}_c^k, \hat{\bf y}') \] 
    }
}

Compute unbiased approximate gradients for $DIV_{\Delta}^{\epsilon}(\Pr\nolimits_c, \Pr\nolimits_c)$ and $DIV_{\Delta}^{\epsilon}(\Pr\nolimits_c, \Pr\nolimits_c)$ as:
\begin{dmath}
	DIV_{\Delta}^{\epsilon}(\Pr\nolimits_p, \Pr\nolimits_c) = \frac{1}{KB} \sum_{k=1}^K \sum_{i=1}^B \Big[ \nabla_{\boldsymbol{\theta}_c} S_k(\hat{\bf y}_a^{(i)}; \boldsymbol{\theta}_c) - \nabla_{\boldsymbol{\theta}_c} S_k(\hat{\bf y}^{(i)}_c; \boldsymbol{\theta}_c) \Big] 
\end{dmath} 
\begin{dmath}
    DIV_{\Delta}^{\epsilon}(\Pr\nolimits_c, \Pr\nolimits_c) =  \frac{2}{K(K-1)B} \sum_{\substack{k,k'=1\\k' \neq k}}^K \sum_{i=1}^B \Big[ \nabla_{\boldsymbol{\theta}_c} S_k(\hat{\bf y}_b^{(i)}; \boldsymbol{\theta}_c) - \nabla_{\boldsymbol{\theta}_c} S_k(\hat{\bf y}_c^{\prime(i)}; \boldsymbol{\theta}_c) \Big]
\end{dmath} 

Update model parameters by descending to the approximated gradients:
\[ \boldsymbol{\theta}_c^{t+1} = \boldsymbol{\theta}_c^{t} - \eta \nabla_{\boldsymbol{\theta}_c}DISC_{\Delta}(\Pr\nolimits_p(\boldsymbol{\theta}_p), \Pr\nolimits_c(\boldsymbol{\theta}_c)) \]

\caption{\emph{Conditional net training algorithm}}
\label{algo1:condNet}
\end{algorithm}

\section{Implementation Details}
\label{ap:implementation_details}
In this section, we provide additional implementation details. For the input pair $({\bf x}, {\bf z}_k)$, the classification branch of the conditional net outputs a score function $\mathcal{G}_k({\bf y}; \boldsymbol{\theta}_c)$, which is a $B \times C$ matrix. We then sample $\hat{\bf y}_c^k$ as described in Section 3.2 of the paper. A non-maximal suppression is applied to further reduce the number of sampled bounding boxes. Corresponding to these samples, we mask the bounding box regression branch of the conditional net such that every bounding box which is not present in the sampled output $\hat{\bf y}_c^k$ is multiplied by a 0 row vector. This ensures that only those bounding boxes which are sampled by the conditional net are retained in the regression branch. The approximated gradients of the loss function is then computed and fed explicitly to the non-differentiable output branch to update the parameters of the network.

\section{Experiments}
\label{ap:experiments}

\def\arraystretch{1.5}
\begin{table*}[t]
\centering
\caption{Detection average precision ($\%$) for different methods on VOC 2012 test set.}
\label{table1:mAP12}
\resizebox{\linewidth}{!}{%
\begin{tabular}{|l|cccccccccccccccccccc|c|}
\hline
\multicolumn{1}{|c|}{Method} & \multicolumn{1}{c}{aero} & \multicolumn{1}{c}{bike} & \multicolumn{1}{c}{bird} & \multicolumn{1}{c}{boat} & \multicolumn{1}{c}{bottle} & \multicolumn{1}{c}{bus} & \multicolumn{1}{c}{car} & \multicolumn{1}{c}{cat} & \multicolumn{1}{c}{chair} & \multicolumn{1}{c}{cow} & \multicolumn{1}{c}{table} & \multicolumn{1}{c}{dog} & \multicolumn{1}{c}{horse} & \multicolumn{1}{c}{mbike} & \multicolumn{1}{c}{pson} & \multicolumn{1}{c}{plant} & \multicolumn{1}{c}{sheep} & \multicolumn{1}{c}{sofa} & \multicolumn{1}{c}{train} & \multicolumn{1}{c}{tv} & \multicolumn{1}{|c|}{mAP} \\ \hline
Jie \emph{et al.}~\cite{Jie_2017_CVPR} & 60.8 & 54.2 & 34.1 & 14.9 & 13.1 & 54.3 & 53.4 & 58.6 & 3.7 & 53.1 & 8.3 & 43.4 & 49.8 & 69.2 & 4.1 & 17.5 & 43.8 & 25.6 & 55 & 50.1 & 38.3 \\
OICR~\cite{Tang_2017_CVPR} & 71.4 & 69.4 & 55.1 & 29.8 & 28.1 & 55.0 & 57.1 & 24.4 & 17.2 & 59.1 & 21.8 & 26.6 & 57.8 & 71.3 & 1.0 & 23.1 & 52.7 & 37.5 & 33.5 & 56.6 & 42.5 \\
W2F~\cite{Zhang_2018_CVPR_W2F} & 73.0 & 69.4 & 45.8 & 30.0 & 28.7 & 58.8 & 58.6 & 56.7 & 20.5 & 58.9 & 10.0 & 69.5 & 67.0 & 73.4 & 7.4 & 24.6 & 48.2 & 46.8 & 50.7 & 58.0 & 47.8 \\
\hline
PredNet (VGG) & 73.1 & 71.4 & 56.3 & 30.8 & 28.7 & 57.6 & 62.1 & 44.6 & 23.4 & 61.7 & 26.4 & 44.4 & 62.7 & 80.0 & 9.1 & 24.4 & 56.8 & 40.2 & 52.8 & 60.8 & \textbf{48.4} \\
PredNet (Ens) & 74.4 & 72.3 & 57.8 & 33.6 & 31.5 & 60.1 & 63.0 & 45.3 & 21.6 & 64.0 & 27.2 & 44.5 & 63.8 & 78.2 & 10.2 & 28.3 & 59.4 & 38.4 & 55.1 & 61.9 & \textbf{49.5} \\
\hline
\end{tabular}
}
\end{table*}

\def\arraystretch{1.5}
\begin{table*}[h]
\centering
\caption{CorLoc (in \%) for different methods on VOC 2012 trainval set.}
\label{table2:corLoc12}
\resizebox{\linewidth}{!}{%
\begin{tabular}{|l|cccccccccccccccccccc|c|}
\hline
\multicolumn{1}{|c|}{Method} & \multicolumn{1}{c}{aero} & \multicolumn{1}{c}{bike} & \multicolumn{1}{c}{bird} & \multicolumn{1}{c}{boat} & \multicolumn{1}{c}{bottle} & \multicolumn{1}{c}{bus} & \multicolumn{1}{c}{car} & \multicolumn{1}{c}{cat} & \multicolumn{1}{c}{chair} & \multicolumn{1}{c}{cow} & \multicolumn{1}{c}{table} & \multicolumn{1}{c}{dog} & \multicolumn{1}{c}{horse} & \multicolumn{1}{c}{mbike} & \multicolumn{1}{c}{pson} & \multicolumn{1}{c}{plant} & \multicolumn{1}{c}{sheep} & \multicolumn{1}{c}{sofa} & \multicolumn{1}{c}{train} & \multicolumn{1}{c}{tv} & \multicolumn{1}{|c|}{mean} \\ \hline
Jie \emph{et al.}~\cite{Jie_2017_CVPR} & 82.4 & 68.1 & 54.5 & 38.9 & 35.9 & 84.7 & 73.1 & 64.8 & 17.1 & 78.3 & 22.5 & 57.0 & 70.8 & 86.6 & 18.7 & 49.7 & 80.7 & 45.3 & 70.1 & 77.3 & 58.8 \\
OICR~\cite{Tang_2017_CVPR} & 89.3 & 86.3 & 75.2 & 57.9 & 53.5 & 84.0 & 79.5 & 35.2 & 47.2 & 87.4 & 43.4 & 43.8 & 77.0 & 91.0 & 10.4 & 60.7 & 86.8 & 55.7 & 62.0 & 84.7 & 65.6 \\
W2F~\cite{Zhang_2018_CVPR_W2F} & 88.8 & 85.8 & 64.9 & 56.0 & 54.3 & 88.1 & 79.1 & 67.8 & 46.5 & 86.1 & 26.7 & 77.7 & 87.2 & 89.7 & 28.5 & 56.9 & 85.6 & 63.7 & 71.3 & 83.0 & 69.4\\
\hline
Pred Net (VGG) & 88.8 & 85.1 & 68.7 & 52.3 & 47.2 & 91.0 & 92.1 & 64.3 & 29.4 & 85.6 & 54.5 & 64.9 & 85.9 & 89.8 & 27.5 & 58.5 & 81.3 & 67.6 & 77.2 & 79.5 & \textbf{69.5} \\
Pred Net (Ens) & 89.1 & 87.1 & 70.3 & 54.2 & 49.8 & 92.5 & 92.5 & 64.6 & 25.1 & 87.0 & 54.8 & 60.5 & 88.3 & 85.4 & 32.6 & 62.7 & 83.4 & 63.2 & 79.9 & 81.7 & \textbf{70.2} \\
\hline
\end{tabular}
}
\end{table*}

\subsection{Ablation Experiments}
In this subsection we discuss the effects of the loss ratio and the thresholding operation on the score function for the detection task on VOC 2007 data set.

\vspace{-4.75mm} 

\paragraph{Effects of the loss ratio:} The loss ratio $\lambda$, as defined in Section~\ref{ssec4.1:TaskLoss} of the main paper, is the ratio of the weight of the localization loss to the weight of the classification loss. In other words, with the higher the loss ratio more importance will be given by the objective function to correctly regress the bounding box labels. We choose three different loss ratios $\lambda = \{1, 0.33, 3\}$ for evaluation. The result of detection task on VOC 2007 test set are $52.1\%$, $51.6\%$ and $52.4\%$ mAP respectively. We empirically observe that assigning more weight to the localization loss helps, indicating that it is important for the networks to tweak the  bounding boxes labels generated from the selective search region proposals. 

\vspace{-4.75mm} 

\paragraph{Effect of thresholding the score function:} As seen in Section~\ref{ssec3.2:probModel} of the main paper, the conditional net generates samples from the score function (\ref{eq4:samplingFromScoringFn}) of the main paper. A low score value indicates that the conditional net is not certain of the bounding box label for an input image. Thresholding the score function would mean that we only sample bounding box labels from the conditional net when it has high certainty over the class distribution. We evaluate the result of the detection task on VOC 2007 test set for the threshold values of $\{0.1, 0.2, 0.3, 0.4, 0.5\}$. Without any threshold, our method has a mean average precision of $51.4\%$. The corresponding mean average precision for the threshold values are $\{51.7\%, 52.2\%, 51.5\%, 51.0\%, 50.6\% \}$. These results indicate that it helps to apply threshold when the network is uncertain over the output classes. This is because we would not like the prediction net to learn from highly uncertain samples. We get the best results for the threshold value of $0.2$. However, we also observe that choosing a large value for threshold has no effect on the detection accuracy. In this case, the network is already reasonably certain of the bounding box label, and we would not like to reject such samples.

Note that for the choice of loss ratio $\lambda = 3$ and threshold kept at $0.2$, our method achieves the best detection average precision of $52.9\%$ mAP.

\subsection{Results on VOC 2012}

Here, we compare our proposed method with other state-of-the art weakly supervised methods. Results for the task of detection average precision (AP) and correct localization (CorLoc) are presented in table~\ref{table1:mAP12} and table~\ref{table2:corLoc12} respectively for PASCAL VOC 2012 data set. Our results are consistent with those observed for VOC 2007 data set and we get an overall increase of 1.7\% over previous state-of-the-art method, W2F~\cite{Zhang_2018_CVPR_W2F}. Our network trained and tested on a single scale outperforms W2F~\cite{Zhang_2018_CVPR_W2F}, which is trained and tested on multiple scales.

\subsection{Results with Region Proposal Networks}

\begin{table}[!h]
\centering
\resizebox{\columnwidth}{!}{%
\begin{tabular}{|c|c|c|c|c|}
\hline
\multicolumn{1}{|c|}{} & \multicolumn{2}{c|}{Selective Search}                        & \multicolumn{2}{c|}{RPN} \\ 
\hline
\multicolumn{1}{|c|}{} & \multicolumn{1}{c|}{mAP \%} & \multicolumn{1}{c|}{CorLoc \%} & \multicolumn{1}{c|}{mAP \%} & \multicolumn{1}{c|}{CorLoc \%} \\ 
\hline
VOC 2007 & 52.9 & 70.9 & 50.9 & 69.1 \\
VOC 2012 & 49.5 & 70.2 & 46.1 & 67.3 \\
\hline
\end{tabular}
}
\caption{Comparison of results when using bounding box proposals from Selective Search and RPN.}
\label{table3:proposalComparison}
\end{table}

\begin{figure*}[!h]
\begin{center}
   \includegraphics[width=0.99\textwidth]{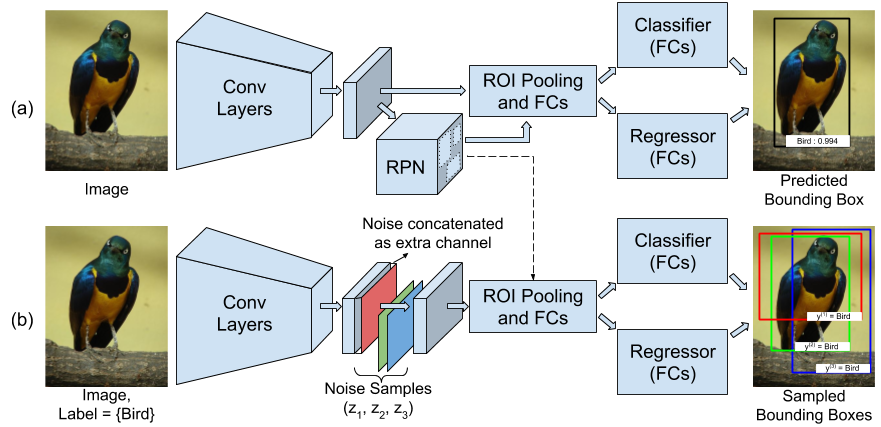} 	
\end{center}
   \caption{\emph{The overall architecture. (a) Prediction Network: a standard Faster-RCNN architecture is used to model the prediction net. For an input image, the region proposal network (RPN) generates a set of bounding box proposals. Features from each of these proposals are computed by the region of interest (ROI) pooling layers, which are then passed through the classifier and regressor to predict the final bounding box. (b) Conditional Network: a modified Fast-RCNN architecture is used to model the conditional net. For a single input image ${\bf x}$ and three different noise samples $\{{\bf z}_1, {\bf z}_2, {\bf z}_3\}$ (represented as red, green and blue matrix), three different bounding boxes $\{{\bf y}^{(1)}, {\bf y}^{(2)}, {\bf y}^{(3)}\}$ are sampled for the given image-level label (bird in this example). Here the noise filter is concatenated as an extra channel to the final convolutional layer. The bounding box proposals required for the conditional net are acquired from the RPN of the prediction net. For both the networks, the initial conv-layers are fixed during training.}}
\label{fig1_supp:Arch}
\end{figure*}

In this subsection we show that our method extends to architectures with region proposal networks (RPN)~\cite{ren2015faster}, thus eliminating the need for external bounding box proposer like Selective Search~\cite{uijlings2013selective}. This enables our framework to perform inference in real-time, while the entire pipeline is trained in an end-to-end fashion including the RPN. 

For this, we replace our prediction net with Faster-RCNN~\cite{ren2015faster} as shown in Figure~\ref{fig1_supp:Arch}. As we wish to use the same set of bounding boxes for both the networks, we share the bounding box proposals generated from RPN as shown in the figure. Furthermore, reusing the computation also makes our training efficient. 

The algorithm proceeds by randomly initializing the RPN and extracting 300 bounding box proposals for each image. These proposals are then fed to the conditional net, which samples the bounding boxes corresponding to the image-level labels for the given image from the proposals. Note that as we introduce noise samples in our conditional net, we get a diverse set of sampled bounding boxes. These bounding boxes are then used to train the conditional net, which also updates the RPN thereby gradually improving the localization of the objects present in the image. 

The results when using bounding box proposals from RPN is presented in Table~\ref{table3:proposalComparison}. We compare the results against those achieved by using Selective Search bounding boxes. Note that, 300 bounding box proposals generated from the randomly initialized RPN has a recall rate of $44.5\% \pm 13.2\%$ on PASCAL VOC 2007 data set. However, after several iterations of training, the final recall rate achieved from 300 bounding box proposals from RPN is $94.7\%$. This is still low when compared to the recall rate achieved by 2000 bounding box proposals from Selective Search method. We argue that due to this difference, we observe a 2\% drop in accuracy. This makes a case of using more bounding box proposals for a better recall rate or using better RPN, like the one proposed in~\cite{Tang_2018_ECCV}. Finally, our choice of employing Faster-RCNN for the prediction net enables our framework to perform inference in real-time.

\clearpage
\clearpage

\end{appendices}

\end{document}